\documentclass[pdflatex,sn-mathphys-num]{sn-jnl}% Math and Physical Sciences Numbered Reference Style
\usepackage{graphicx}%
\usepackage{multirow}%
\usepackage{amsmath,amssymb,amsfonts,amsxtra,bm,bbm,amssymb}%
\usepackage{amsthm}%
\usepackage{mathrsfs}%
\usepackage[title]{appendix}%
\usepackage{xcolor}%
\usepackage{textcomp}%
\usepackage{manyfoot}%
\usepackage{booktabs}%
\usepackage{algorithm}%
\usepackage{algorithmicx}%
\usepackage{algpseudocode}%
\usepackage{listings}%
\usepackage{pifont}
\usepackage{microtype}
\usepackage{mathtools}
\DeclareMathOperator{\support}{supp}

\clearpage{}%

\newcommand{\Dc}{\mathcal{D}}

\newcommand{\Pc}{\mathcal{P}}

\newcommand{\Oc}{\mathcal{O}}
\newcommand{\hmu}{\hat{\mu}}

\newcommand{\E}{\mathbb{E}}

\newcommand{\reals}{\mathbb{R}}

\newcommand{\ip}[2]{\langle {#1},\, {#2} \rangle}

\newcommand{\hDc}{\hat{\mathcal{D}}}
\newcommand{\lpt}{-\log P_\theta(y|x)}
\newcommand{\pt}{\log P_\theta(y|x)}
\newcommand{\supp}[1]{\support\left({#1}\right)}

\newtheorem{assumption}{Assumption}
\theoremstyle{thmstyleone}%
\newtheorem{theorem}{Theorem}%  meant for continuous numbers
\theoremstyle{thmstyletwo}%

\theoremstyle{thmstylethree}%

\begin{document}

\title[Fundamental Safety-Capability Trade-offs in Fine-tuning Large Language Models]{Fundamental Safety-Capability Trade-offs in Fine-tuning Large Language Models}

%%=============================================================%%
%% GivenName	-> \fnm{Joergen W.}
%% Particle	-> \spfx{van der} -> surname prefix
%% FamilyName	-> \sur{Ploeg}
%% Suffix	-> \sfx{IV}
%% \author*[1,2]{\fnm{Joergen W.} \spfx{van der} \sur{Ploeg} 
%%  \sfx{IV}}\email{iauthor@gmail.com}
%%=============================================================%%

\author*[1]{\fnm{Pin-Yu} \sur{Chen}}\email{pin-yu.chen@ibm.com}\equalcont{These authors contributed equally to this work.}

\author[2]{\fnm{Han} \sur{Shen}}\email{shenh5@rpi.edu}
\equalcont{These authors contributed equally to this work.}

\author[1]{\fnm{Payel} \sur{Das}}\email{daspa@us.ibm.com}

\author[2]{\fnm{Tianyi} \sur{Chen}}\email{chent18@rpi.edu}

\affil*[1]{\orgname{IBM Research}, \orgaddress{\street{1101 Kitchawan Road}, \city{Yorktown Heights}, \postcode{10601}, \state{New York}, \country{USA}}}

\affil[2]{\orgdiv{Department of Electrical, Computer, and Systems Engineering}, \orgname{Rensselaer Polytechnic Institute}, \orgaddress{\street{Jonsson Engineering Center 110 8th Street}, \city{Troy}, \postcode{12180}, \state{New York}, \country{USA}}}

%%==================================%%
%% Sample for unstructured abstract %%
%%==================================%%

\abstract{Fine-tuning Large Language Models (LLMs) on some task-specific datasets has been a primary use of LLMs. However, it has been empirically observed that this approach to enhancing capability  inevitably compromises safety, a phenomenon also known as the safety-capability trade-off in LLM fine-tuning. This paper presents a theoretical framework for understanding the interplay between safety and capability in two primary safety-aware LLM fine-tuning strategies, providing new insights into the effects of
data similarity, context overlap, and alignment loss landscape. Our theoretical results characterize the fundamental limits of the safety-capability trade-off in LLM fine-tuning, which are also validated by numerical experiments.}


%%\pacs[JEL Classification]{D8, H51}

%%\pacs[MSC Classification]{35A01, 65L10, 65L12, 65L20, 65L70}

\maketitle

\section{Introduction}\label{sec1}

Large language models (LLMs) are transformer-based neural networks that are pre-trained on large textual datasets using next-token prediction loss and further refined to follow instructions and achieve compliance. The latter process is also known as \textit{alignment}, where machine learning techniques such as supervised fine-tuning (SFT) and reinforcement learning with human feedback \cite{ouyang2022training} are used to update the pre-trained model weights to align the LLM's response with the desired output. In particular, safety alignment focuses on preventing LLMs from generating harmful responses, by teaching LLMs to refuse to answer unsafe user queries.
Beyond the alignment stage, an aligned LLM is often fine-tuned on a domain-specific dataset to improve its capabilities in downstream tasks \cite{ding2023parameter}, such as coding, reasoning, and mathematical problem-solving. However, recent studies have found that the increase in capability comes at a hidden cost of breaking the innate safety guardrail, even when the task-specific fine-tuning dataset does not contain any malicious data samples \cite{qi2024finetuning}. Such an unwanted degradation of safety after fine-tuning poses significant challenges to the usability and reliability of LLMs \cite{chen2025computational,choudhury2025promise,chang2025red}.

In this paper, we establish a theoretical framework to study the problem of safety-capability trade-off in fine-tuning LLMs. While recent work has provided empirical evidence of safety degradation after LLM fine-tuning and proposed various mitigation methods \cite{huang2024harmful},
a comprehensive theoretical understanding of the interplay between safety and capability in LLM fine-tuning remains elusive. 
See Section \ref{appen:illustration} for a motivating illustration of our problem setup. 
To fill this critical gap, we provide theoretical interpretations and novel insights to characterize the safety-capability trade-offs in LLM fine-tuning. Our framework considers two practical safety-aware fine-tuning strategies: (i) \textit{Alignment Loss Constraint} -- where a proxy safety instruction-tuning dataset is used together with the downstream capability dataset during fine-tuning to constrain the safety loss, such as    \cite{bianchi2024safetytuned,shen2025seal}; and, (ii) \textit{Alignment Parameter Constraint} -- where the model parameter updates are constrained to be in a local neighborhood of the aligned model to maintain a similar level of safety after fine-tuning, such as \cite{peng2024navigating,hsu2024safe}. 

We summarize our key findings as follows.
\begin{itemize}
    \item For the alignment loss constraint strategy, higher similarity between the original and proxy safety data provably mitigates safety degradation. In addition, less context overlap between the safety and capability data provably improves the safety-capability trade-off. 
    \item For the alignment parameter constraint strategy, the sensitivity of the local landscape around the originally aligned LLM in the model parameter space is shown to control the safety-capability trade-off. 
    \item Numerical experiments on LLMs validate our theoretical findings to characterize the effects of data similarity and context overlap on the safety-capability trade-off.  
\end{itemize}

\section{Results}\label{sec2}

\subsection{Overview and Mathematical Notations} 
We first provide the mathematical notations, formalize the theoretical analysis of safety-aware LLM fine-tuning, and then present numerical results to validate our findings. All proofs are given in the Methods section.

Let $\mathcal{V}$ be a finite token set of an LLM. Define $x\in\mathcal{V}^{d_x}/y\in\mathcal{V}^{d_y}$ as an input/output token sequence, where $d_x$ and $d_y$ are their token lengths. Safety alignment refers to instruction tuning on a set of inputs and their preferred outputs, $\{x,y\}$.
Define $\Dc_s(x)$ and $\mu_s(y|x)$ respectively as the original safety alignment input distribution and the target output distribution, which might be inaccessible in the fine-tuning process (e.g., the alignment dataset was not released). Instead, one may use a proxy safety instruction-tuning dataset sampled from the alternative input and output distributions for fine-tuning, denoted as $\hDc$ and $\hmu(\cdot|x)$, respectively.
We also denote $\Dc_f(x)$ and $\mu_f(y|x)$ as the input and output distributions of the downstream task data.
The goal is to fine-tune the model on $\Dc_f(x)$ and  $\mu_f(y|x)$  for capability improvement while preserving the model's safety alignment on $\Dc_s$ and $\mu_s$. For brevity, given $x$, we write $\mu_s(\cdot|x)$ as $\mu_s(x)$ and similarly for other output distributions. We use $\Oc$ for the big $O$ notation, $\|\cdot \|$ for a canonical norm, $\|\cdot\|_{TV}$ for the total variation, $D_{KL}(\cdot|\cdot)$ for the Kullback–Leibler divergence, $\supp \cdot$ for the support of a probability distribution, $\nabla$ for gradient, and $\ip{\cdot}{\cdot}$ for inner product.

\subsection{Case I: Alignment Loss Constraint}
In the case where $\Dc_s$ and $\mu_s(\cdot|x)$ are unknown, the fine-tuner has access to proxy distributions  $\hDc$ and $\hmu(\cdot|x)$ to measure the safety performance. Using both the task and safety data, we formulate LLM fine-tuning as the following problem:
%\begin{small}
\begin{align}\label{formulation:safety loss constraint}
    &\min_{\theta\in\Theta} \E_{x\sim\Dc_f,y\sim\mu_f(x)}[-\log P_\theta(y|x)],
    \\ \nonumber 
    &~{\rm s.t.}~~\E_{x\sim\hDc,y\sim\hmu(x)}[-\log P_\theta(y|x)] \leq \epsilon_1
\end{align}
%\end{small}%
where $\theta$ is the set of LLM's trainable parameters, $\Theta$ is a closed convex constraint set, and $P_\theta(y|x)$ is the model's probability of outputting $y$ given input $x$.
We use the penalty method to solve the constraint optimization problem:
%\begin{small}
\begin{align}\label{formulation:safety loss penalty}
    \min_{\theta\in\Theta} \E_{x\sim\Dc_f,y\sim\mu_f(x)}[-\log P_\theta(y|x)]+\lambda \cdot \E_{x\sim\hDc,y\sim\hmu(x)}[-\log P_\theta(y|x)]
\end{align}
%\end{small}%
where $\lambda \geq 0$ is the coefficient (strength) of the penalty term.

To quantify how much safety alignment is compromised, we define the safety alignment gap of model $\Pc_\theta(y|x)$ between the input distribution $\Dc_s$ and the target distribution $\mu_s$ as
%\begin{small}
\begin{align}\label{eq:safety alignmetn gap}
    G_s(P_\theta)\coloneqq&\E_{x\sim\Dc_s,y\sim\mu_s(x)}[-\log P_\theta(y|x)] -\E_{x\sim\Dc_s,y\sim\mu_s(x)}[-\log \mu_s(y|x)]
\end{align}
%\end{small}%
where $\E_{x\sim\Dc_s,y\sim\mu_s(x)}[-\log \mu_s(y|x)]$ is essentially the minimum of $\min_{P}\E_{x\sim\Dc_s,y\sim\mu_s(x)}[-\log P(y|x)]$. A smaller $G_s(P_\theta)$ suggests a better safety alignment after fine-tuning.
We make the following assumptions for Case I.

\begin{small}
\begin{assumption}[Bounded log probability]\label{assumption:bounded log probability}
    Assume for any $\theta\in\Theta$, we have $\log P_\theta(y|x) \leq C_p$ given any $x$ and $y$.
\end{assumption}
\end{small}

\begin{small}
\begin{assumption}[Realizable output distribution]\label{assumption:realizable target}
    Assume the output distributions $\mu_s,\hmu$, and $\mu_f$ are realizable by the parameterization $P_\theta$, that is, $\mu_s(x),\hmu(x)$, and $\mu_f(x)$ belong to $\{P_\theta(x):\theta\in\Theta\}$ given any $x$.
\end{assumption}
\end{small}
Assumptions \ref{assumption:bounded log probability} and \ref{assumption:realizable target} mean the considered LLM family should be able to learn the desired output distributions.
Our first result is on the safety alignment guarantee.
\begin{theorem}[Safety alignment loss gap in Case I]\label{theorem:alignment loss constraint alignment bound}
Under Assumptions \ref{assumption:bounded log probability} and \ref{assumption:realizable target}, any solution of \eqref{formulation:safety loss penalty} denoted as $\theta$ satisfies the following safety alignment guarantee:
    %\begin{small}
    \begin{align}
         G_s(P_\theta)
        \!&=\!\Oc\Big(\frac{1}{\lambda}\Big)\!+\!\Oc\Big(\big\|\hDc-\Dc_s\big\|_{TV}\!\Big)\!+\!\Oc\Big(\E_{x\sim\Dc_s}\big\|\mu_s(x)-\hmu(x)\big\|_{TV}\!\Big) \nonumber \\
&~~~~+\Oc\Big(\!\E_{x\sim\Dc_s}\big[D_{KL}\big(\mu_s(x)|\hmu(x)\big)\big]\!\Big).\nonumber
    \end{align}
    %\end{small}
\end{theorem}
% \begin{proof}
% %Please see Appendix \ref{proof_theorem:alignment loss constraint alignment bound} in Supporting Information. 
% \end{proof}
The first term quantifies the influence of the regularization coefficient $\lambda$ on the safety alignment gap. The second term characterizes the effect of distribution mismatch between the original and proxy input distributions. The last two terms specify the impact of distribution mismatch between the original and proxy output distributions.

Next, similar to \eqref{eq:safety alignmetn gap}, we define the capability performance gap of model $P_\theta$ on $\Dc_f$ and $\mu_f$ as
%\begin{small}
\begin{align}
    \label{eq:capability gap}
    G_f(P_\theta)\coloneqq &\E_{x\sim\Dc_f,y\sim\mu_f(x)}[\lpt] -\E_{x\sim\Dc_f,y\sim\mu_f(x)}[-\log \mu_f(y|x)].
\end{align}
%\end{small}%
A smaller $G_f(P_\theta)$ suggests a stronger capability after fine-tuning, and $G_f(P_\theta)$ is upper-bounded as follows.

\begin{theorem}[Capability loss gap in Case I]\label{theorem:alignment loss constraint ft bound}
   Assume Assumptions \ref{assumption:bounded log probability} and \ref{assumption:realizable target} hold. Then any solution $\theta$ of \eqref{formulation:safety loss penalty} satisfies the following fine-tuning guarantee:
%   \begin{small}
    \begin{align}
        G_f(P_\theta) \leq \lambda \sum_{x\in\supp{\hDc} \cap \supp{\Dc_f}}\hDc(x)D_{KL}(\hmu(x)|\mu_f(x)).\nonumber
    \end{align}
    %\end{small}
\end{theorem}

% \begin{proof}
% Please see Appendix \ref{proof_theorem:alignment loss constraint ft bound} in Supporting Information. 
% \end{proof}
Theorem \ref{theorem:alignment loss constraint ft bound} reveals several new insights into how fine-tuning with a safety alignment loss constraint conflicts with the capability performance. Enlarging the penalty strength $\lambda$ increases the upper bound of $G_f(P_\theta)$ because the fine-tuning process puts more emphasis on safety alignment. Moreover, suppose $\hDc$ and $\Dc_f$ have notable overlap in their support (which we call the \textit{context overlap}). In that case, the bound can be increased due to direct conflicts between safety and capability objectives, especially when their output distributions $\hmu(x)$ and $\mu_f(x)$ are divergent.

\subsection{Case II: Alignment Parameter Constraint}
This case constrains the updates of an aligned model  $\theta_s$ to its local neighborhood in the model parameter space during fine-tuning, with the premise that the fine-tuned model would maintain similar safety alignment to $\theta_s$. With the downstream data sampled from $\Dc_f$ and $\mu_f$,
we aim to solve the following problem:
\begin{align}\label{formulation:parameter constraint}
    \min_{\theta\in\reals^d} \E_{x\sim\Dc_f,y\sim\mu_f(x)}[-\log P_\theta(y|x)],~{\rm s.t.}~~\|\theta-\theta_s\| \leq \epsilon_2.
\end{align}
We also note that this fine-tuning strategy does not require additional safety data.
We make the following assumptions.
\begin{small}
\begin{assumption}[Local Lipschitz continuity]\label{assumption:locally continuous param}
    % Given any $\theta\in\reals^d$ and $\epsilon>0$, assume there exists $L_s(\theta,\epsilon)>0$ such that for any $\theta_1,\theta_2$ satisfying $\|\theta_1-\theta\|\leq \epsilon$ and $\|\theta_2-\theta\|\leq \epsilon$, it holds that
     Given $\theta_s$ and $\epsilon_2$, assume there exists $L_s(\theta_s,\epsilon_2)>0$ such that for any $\theta$ satisfying $\|\theta-\theta_s\|\leq \epsilon_2$, it holds that
    % \begin{align}
    %     \big\|\E_{x\sim\Dc_s,y\sim\mu_s(x)}[\log P_{\theta_1}(y|x)]-\E_{x\sim\Dc_s,y\sim\mu_s(x)}[\log P_{\theta_2}(y|x)]\big\| \leq L_s(\theta,\epsilon) \|\theta_1-\theta_2\|^2
    % \end{align}
    %\begin{small}
    \begin{align}
        &\big\|\E_{x\sim\Dc_s,y\sim\mu_s(x)}[\log P_{\theta}(y|x)]-\E_{x\sim\Dc_s,y\sim\mu_s(x)}[\log P_{\theta_s}(y|x)]\big\| \leq L_s(\theta_s,\epsilon_2) \|\theta-\theta_s\|. \nonumber
    \end{align}
    %\end{small}
    % and similar condition holds for $\E_{x\sim\Dc_f,y\sim\mu_f(x)}[\nabla\log P_{\theta_1}(y|x)]$ with $L_f(\theta,\epsilon)$.
\end{assumption}
\end{small}
Using the same notion of safety and capability gaps as Case I, we first have the safety alignment guarantee:
\begin{theorem}[Safety alignment loss gap in Case II]\label{theorem:alignment param constraint alignment bound}
    Assume Assumption \ref{assumption:locally continuous param} holds. Then any $\theta$ which is the solution of \eqref{formulation:parameter constraint} satisfies
    %\begin{small}
    \begin{align}
        G_s(P_\theta)\leq L_s(\theta_s,\epsilon_2)\epsilon_2+G_s(P_{\theta_s}). \nonumber
    \end{align}
    %\end{small}%
    where $G_s(P_{\theta_s})$ is the safety alignment gap of the  model $\theta_s$.
\end{theorem}
% \begin{proof}
% Please see Appendix \ref{proof_theorem:alignment param constraint alignment bound} in Supporting Information. 
% \end{proof}
Theorem \ref{theorem:alignment param constraint alignment bound} shows how the changes of the loss landscape around $\theta_s$, captured by the local Lipschitz constraint $L_s$ and the neighborhood range $\epsilon_2$, affect the safety gap. To study the capability under Case II, we make an additional assumption on the local smoothness of the aligned model parameters.

\begin{small}
\begin{assumption}[Local Lipschitz smoothness]\label{assumption:locally_smooth_param}
    Given $\theta_s$ and $\epsilon_2$, assume there exists $L_f'(\theta_s,\epsilon_2)>0$ such that for any $\theta$ satisfying $\|\theta-\theta_s\|\leq \epsilon_s$, it holds that
    % \begin{align}
    %     \big\|\E_{x\sim\Dc_s,y\sim\mu_s(x)}[\nabla\log P_{\theta_1}(y|x)]-\E_{x\sim\Dc_s,y\sim\mu_s(x)}[\nabla\log P_{\theta_2}(y|x)]\big\| \leq L_s'(\theta,\epsilon) \|\theta_1-\theta_2\|^2
    % \end{align}
    %\begin{small}
    \begin{align}
        &\E_{x\sim\Dc_f,y\sim\mu_f(x)}[-\log P_{\theta}(y|x)]-\E_{x\sim\Dc_f,y\sim\mu_f(x)}[-\log P_{\theta_s}(y|x)]\nonumber\\
    &\leq \ip{-\E_{x\sim\Dc_f,y\sim\mu_f(x)}[\nabla_{\theta}\log P_{\theta_s}(y|x)]}{\theta-\theta_s}+\frac{L_f'(\theta_s,\epsilon_2)}{2}\|\theta_s-\theta\|^2. \nonumber
    \end{align}
    %\end{small}
    % and similar condition holds for $\E_{x\sim\Dc_f,y\sim\mu_f(x)}[\nabla\log P_{\theta_1}(y|x)]$ with $L_f'(\theta,\epsilon)$.
\end{assumption}
\end{small}

With Assumption \ref{assumption:locally_smooth_param}, we obtain an upper bound on the capability gap of a fine-tuned model $\theta$ under Case II.
\begin{theorem}[Capability loss gap in Case II]\label{theorem:alignment param constraint ft bound}
    Under Assumption \ref{assumption:locally_smooth_param}, any solution of \eqref{formulation:parameter constraint} denoted as $\theta$ satisfies
        %\begin{small}
    \begin{equation}
        G_f(P_\theta) \leq -\frac{\big\| \E_{x\sim\Dc_f,y\sim\mu_f(x)}[\nabla_{\theta}\log P_{\theta_s}(y|x)]\big\|^2}{2 L_f'(\theta_s,\epsilon_2)}+G_f(P_{\theta_s}). \nonumber
    \end{equation}
           %\end{small}
\end{theorem}
% \begin{proof}
% Please see Appendix \ref{proof_theorem:alignment param constraint ft bound} in Supporting Information. 
% \end{proof}
Theorem \ref{theorem:alignment param constraint ft bound} shows the capability gap is governed by the sensitivity of original aligned LLM $\theta_s$ on task data, measured by the gradient norm and the smoothness constant $L'_f$.

\subsection{Experiments}

We design numerical experiments using Llama-2-7B base model \citep{touvron2023llama}.
The instruction-tuning datasets used are: 1) Orca \citep{OpenOrca}. The inputs are various instructions covering summarization tasks, reasoning tasks, etc. The target outputs are generated by GPT-4 \citep{achiam2023gpt}; 2) Alpaca/Alpaca-GPT-4  \citep{alpaca}. The inputs are similar to Orca. The target outputs are generated by OpenAI's text-davinci-003 or GPT-4; 3) Commitpackft \citep{muennighoff2023octopack}. The inputs are GitHub commits formatted into code modification requests, and the target outputs are modified codes after the commits; 4) Open-platypus \citep{platypus2023}. The inputs are reasoning problems, and the target outputs are generated by GPT-4 or by humans. 

Before fine-tuning, we first align the Llama-2-7B base model on Orca, which is treated as the alignment dataset generated by $\Dc_s$ and $\mu_s$. To validate our theoretical analysis, we measure the safety alignment loss gap $G_s(P_\theta)$ in \eqref{eq:safety alignmetn gap} and the capability loss gap $G_f(P_\theta)$ in \eqref{eq:capability gap}
by the empirical loss of the associated data samples. 
%The expectation on $\Dc_s,\mu_s$ or $\Dc_f,\mu_f$ are approximated with empirical surrogate on the samples from the alignment dataset or fine-tuning dataset. 
Additionally, the minimum value terms $\E_{x\sim\Dc_s,y\sim\mu_s(x)}[-\log \mu_s(y|x)]$ or $\E_{x\sim\Dc_f,y\sim\mu_f(x)}[-\log \mu_f(y|x)]$ are approximated with the Llama-2-7B based model trained on the alignment dataset or the fine-tuning dataset without any constraints.

\begin{figure}[t]
    \centering
    \includegraphics[width=0.49\linewidth]{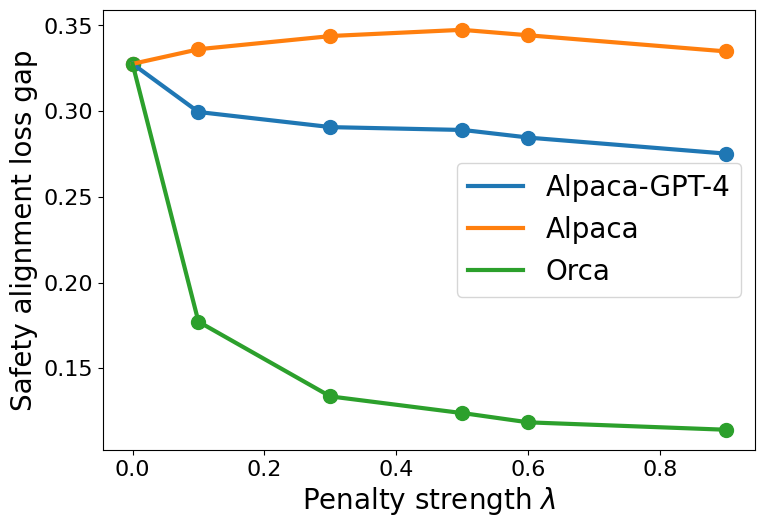}
    \includegraphics[width=0.49\linewidth]{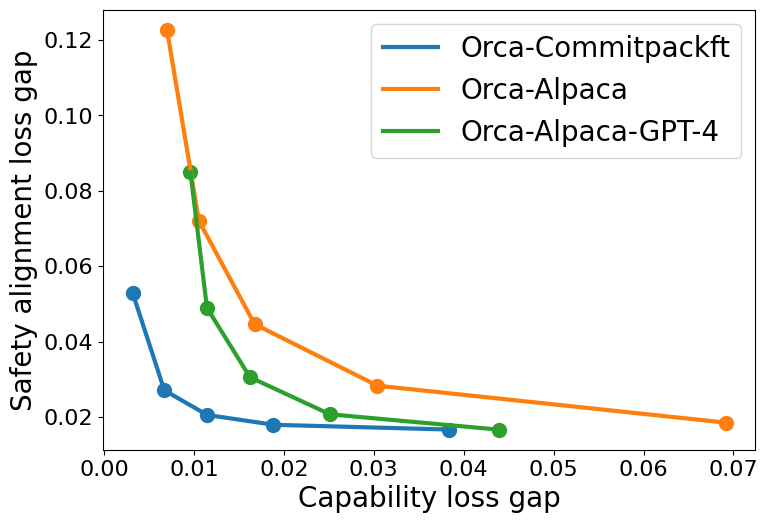}
    %\vspace{-7mm}
    \caption{\textit{Left}: Alignment loss gap in Case I with varying penalty strength $\lambda$ and different proxy alignment datasets (indicated by the legend). \textit{Right}: Safety-capable trade-off in Case I. The legend indicates [alignment dataset]-[fine-tuning dataset].}
    \label{fig:loss constraint}
        %\vspace{-4mm}
\end{figure}

%\paragraph{Main Findings.}
We elaborate on our main findings as follows.

\textbf{Higher similarity between the original and proxy safety alignment datasets is better at reducing the safety alignment gap in Case I.}
%Theorem \ref{theorem:alignment loss constraint alignment bound} states that the similarity of the original and proxy alignment datasets affects the safety alignment loss gap. 
To validate the similarity analysis in Theorem \ref{theorem:alignment loss constraint alignment bound},
we vary the penalty coefficient $\lambda$ in \eqref{formulation:safety loss penalty} with different proxy alignment datasets and report the results in Figure \ref{fig:loss constraint} (Left).
Note that Alpaca-GPT-4 and Alpaca have the same inputs, and thus the same $\hDc$.
We observe that for any $\lambda>0$, using Alpaca-GPT-4 (blue curve) as the proxy alignment dataset achieves a lower alignment loss gap than using Alpaca (orange curve). This is because the target outputs of Alpaca-GPT-4 and Orca are both generated by GPT-4, while Alpaca's target outputs are generated by text-davinci-003. As a sanity check, when Orca (green curve) is also used as the proxy alignment dataset, the alignment loss gap can be further decreased, because $\hDc=\Dc_s$ and $\hDc = \Dc_s$.

\textbf{The context overlap between alignment and capability datasets controls the safety-capability trade-off in Case I.} 
We examine the influence of different 
fine-tuning datasets on capability under Case I, as characterized in Theorem \ref{theorem:alignment loss constraint ft bound}. We use Orca as the proxy alignment dataset. Figure \ref{fig:loss constraint} (Right) shows the safety-capability trade-off curve by setting $\lambda=\{0.1, 0.3, 0.5, 0.7, 0.9\}$, where increased capability (smaller fine-tuning loss gap) comes at the cost of decreased safety (larger alignment loss gap), as also indicated by our theoretical analysis. Theorem \ref{theorem:alignment loss constraint alignment bound} suggests that the alignment loss gap decreases with $\lambda$, while the term $\sum \hDc(x)D_{KL}(\mu_s(x)|\mu_f(x))$ in Theorem \ref{theorem:alignment loss constraint ft bound} increases with $\lambda$ when the fine-tuning loss puts more emphasis on alignment.

\begin{figure}[t]
    \centering
    \includegraphics[width=0.49\linewidth]{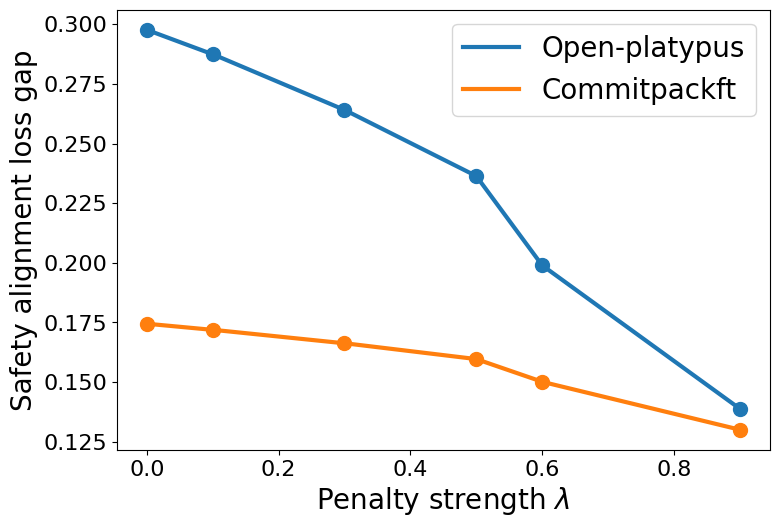}
    \includegraphics[width=0.49\linewidth]{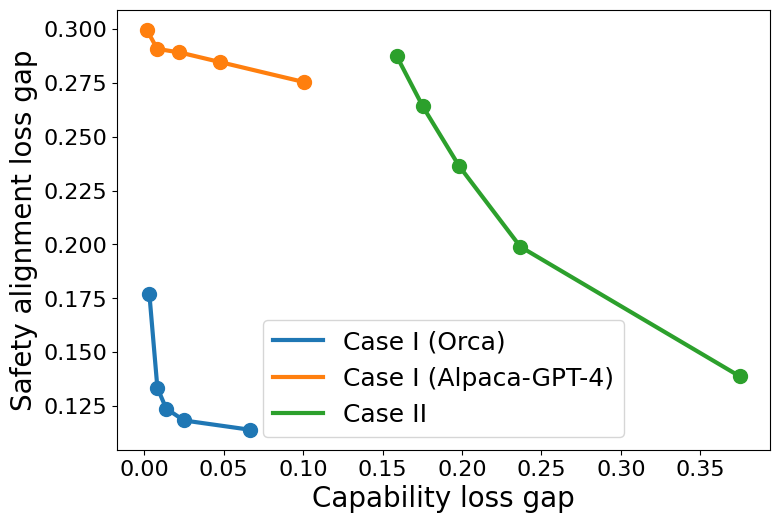}
        %\vspace{-7mm}
    \caption{ \textit{Left}: Alignment loss gap in Case II with varying penalty strength $\lambda$ and different task datasets (indicated by the legend). 
    \textit{Right}: Safety-capability comparison of Case I and Case II. }
    \label{fig:param constraint and comparison}
        %\vspace{-4mm}
\end{figure}

%The first observation is that for all curves, there exists a trade-off between alignment and fine-tuning performance, which is consistent with our theoretical bounds. In Theorem \ref{theorem:alignment loss constraint alignment bound}, we have the alignment loss gap is decreasing with $\lambda$ since there is no alignment approximation in this test. Then in Theorem \ref{theorem:alignment loss constraint ft bound}, since the term $\sum_x \hDc(x)D_{KL}(\mu_s(x)|\mu_f(x))$ is non-zero in this test, we know the fine-tuning loss gap increasing with $\lambda$. Therefore, there exists an internal trade-off between alignment and fine-tuning in this test, which is fully controlled by $\lambda$ as demonstrated in the figure. 
Theorem \ref{theorem:alignment loss constraint ft bound} also explains why fine-tuning on Commitpackft (blue curve) achieves the smallest capability loss gap, because Commitpackft's inputs focus on code generation prompts, which have little context overlap with Orca's input domain (general text). On the other hand, Alpaca and Orca have similar input domains, and thus a larger intersected set $\supp{\Dc_f}\cap \supp{\hDc}$ leads to a larger capability loss.

\textbf{Case II is more restrictive than Case I for capability improvement.} 
%To study the effect of the alignment parameter constraint, 
We solve \eqref{formulation:parameter constraint} by penalizing the parameter constraint function $\|\theta_s-\theta\|^2$ onto the objective and minimize $\E_{\Dc_f,\mu_f}[\lpt]+\lambda \cdot \|\theta-\theta_s\|^2$. Therefore, decreasing $\epsilon_2$ is effectively increasing $\lambda$. Figure \ref{fig:param constraint and comparison} (Left) verifies Theorem \ref{theorem:alignment param constraint alignment bound} that a smaller $\epsilon_2$ leads to a lower alignment loss gap for different fine-tuning datasets. Finally, Figure \ref{fig:param constraint and comparison} (Right) compares the safety-capability trade-offs for Case I and Case II. While Case II can achieve a small alignment loss gap, the local neighborhood constraint imposed when fine-tuning the model parameters limits the capability improvement, resulting in a larger capability loss gap than Case I.

\section{Conclusion}\label{sec13}

This paper established a theoretical analysis to understand the fundamental trade-offs between safety and capability when fine-tuning LLMs. Our theoretical and empirical results unveil how data similarity and context overlap affect safety and capability. The insights from our findings can inform future research and practice of AI safety in LLMs.

\section{Methods}\label{sec11}

\subsection{Problem Illustration}
\label{appen:illustration}

\begin{figure}[t]
    \centering
    \includegraphics[width=0.95\linewidth]{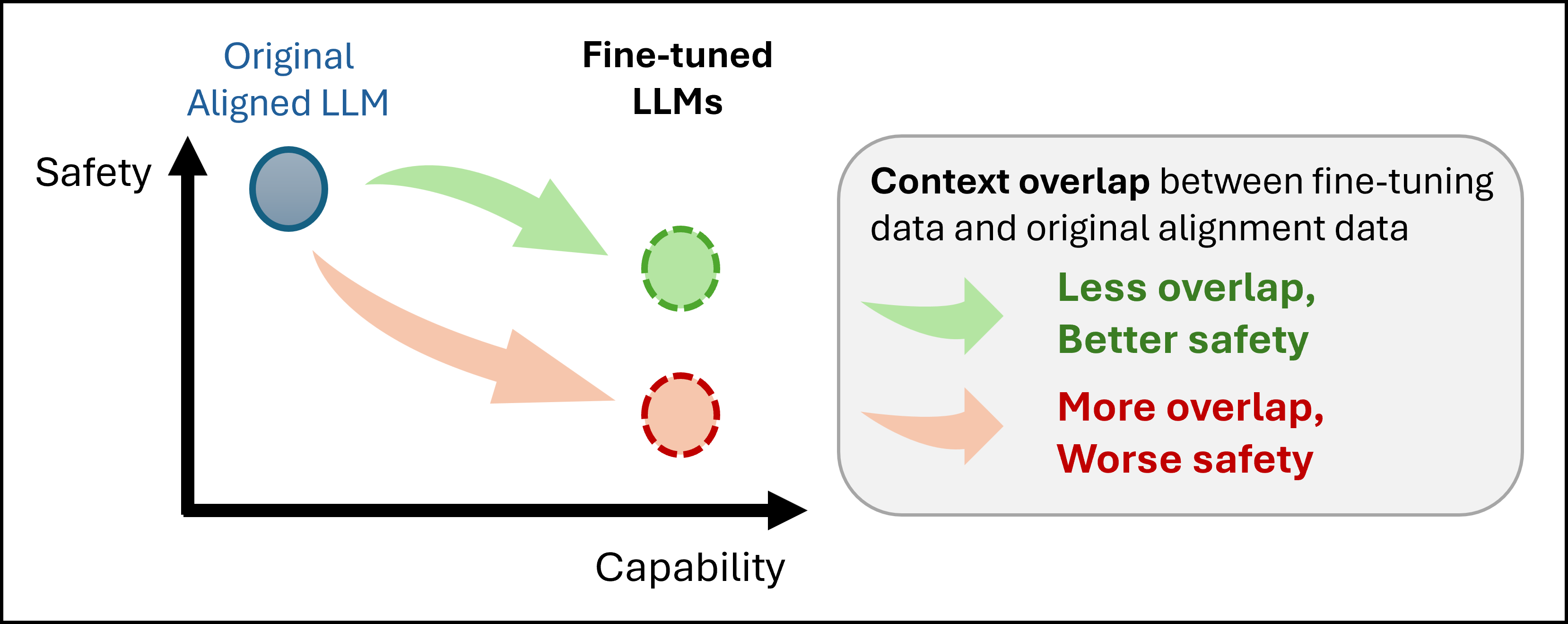}
    \caption{Illustration of safety-capability trade-offs in LLM fine-tuning concerning context overlap. Here, context overlap refers to the overlap of input distributions between the proxy alignment data and the fine-tuning data in the case of alignment loss constraint (Case I).}
    \label{fig:motivation}
\end{figure}

Figure \ref{fig:motivation} illustrates the motivation for studying the trade-offs between safety and capability in LLM fine-tuning. In particular, we prove theoretically and empirically that the context overlap between fine-tuning data and original alignment data plays an important role in the safety of fine-tuned LLMs. More context overlap leads to less safety after LLM fine-tuning.

\subsection{Proof of Theorem \ref{theorem:alignment loss constraint alignment bound}}
\label{proof_theorem:alignment loss constraint alignment bound}
In this proof, we will write $\E_{x\sim\Dc,y\sim\mu(x)}[-\log P_\theta(y|x)]$ as $\E_{\Dc,\mu}[-\log P_\theta(y|x)]$ for brevity. We first decompose the safety alignment gap as:
    \begin{align}\label{eq:idk0}
      G_s(P_\theta):=   &\E_{x\sim\Dc_s,y\sim\mu_s(x)}[-\log P_\theta(y|x)]-\E_{x\sim\Dc_s,y\sim\mu_s(x)}[-\log \mu_s(y|x)] \nonumber\\
        =& \E_{\Dc_s,\mu_s}[\lpt]-\E_{\hDc,\hmu}[\lpt]+\E_{\hDc,\hmu}[\lpt]+\E_{\hat{\Dc},\hat{\mu}}[\log\hat{\mu}(y|x)]\nonumber\\
        &+\E_{\hDc,\hmu}[-\log \hat{\mu}(y|x)]-\E_{\Dc_s,\mu_s}[-\log \mu_s(y|x)] .
        % &\leq \E_{\Dc_s,\mu_s}[\lpt]\!-\!\E_{\hDc,\hmu}[\lpt]\!+\!\frac{C_p}{\lambda}\!+\!\E_{\hDc,\hmu}[\log \hat{\mu}(y|x)]\!-\!\E_{\Dc_s,\mu_s}[-\log \mu_s(y|x)]
    \end{align}
    The first difference in \eqref{eq:idk0} can be bounded as
    \begin{align}\label{eq:idk4}
        &\E_{\Dc_s,\mu_s}[\lpt]-\E_{\hDc,\hmu}[\lpt] \\ \nonumber
        &= -\sum_x\sum_y \big(\hDc(x)\hmu(y|x)-\Dc_s(x)\mu_s(y|x)\big)\pt \nonumber\\
        &\leq C_p \sum_x\sum_y \big|\hDc(x)\hmu(y|x)-\Dc_s(x)\mu_s(y|x)\big|\nonumber\\
        &\leq C_p \sum_x |\hDc(x)-\Dc_s(x)|+C_p\sum_x \Dc_s(x)\sum_y|\hmu(y|x)-\mu_s(y|x)|
    \end{align}
    where we used Assumption \ref{assumption:bounded log probability} in the first inequality.
    
    Next we bound the second difference in \eqref{eq:idk0}. By the optimality of $\theta$ for \eqref{formulation:safety loss penalty}, we have
    \begin{align}
        \E_{\Dc_f,\mu_f}[\lpt]+\lambda\E_{\hat{\Dc},\hat{\mu}}[-\log P_\theta(y|x)]\leq  \E_{\Dc_f,\mu_f}[-\log \hmu(y|x)]+\lambda\E_{\hat{\Dc},\hat{\mu}}[-\log \hmu(y|x)].
    \end{align}
    Rearranging the above inequality and dividing both sides by $\lambda$ yields
    \begin{align}\label{eq:second diff}
        \E_{\hat{\Dc},\hat{\mu}}[-\log P_\theta(y|x)]+\E_{\hat{\Dc},\hat{\mu}}[\log \hmu(y|x)] 
        &\leq \frac{1}{\lambda}\Big(\E_{\Dc_f,\mu_f}[-\log \hmu(y|x)]-\E_{\Dc_f,\mu_f}[\lpt]\Big) \nonumber\\
        &\leq \frac{2 C_p}{\lambda}
    \end{align}
    where the last inequality follows from Assumptions \ref{assumption:bounded log probability} and \ref{assumption:realizable target}.
    The last difference in \eqref{eq:idk0} can be further decomposed into
    \begin{align}
        &\E_{\hDc,\hmu}[-\log \hat{\mu}(y|x)]-\E_{\Dc_s,\mu_s}[-\log \mu_s(y|x)] \nonumber\\
        &=\E_{\hDc,\hmu}[-\log \hat{\mu}(y|x)]-\E_{\Dc_s,\mu_s}[-\log \hat{\mu}(y|x)]+\E_{\Dc_s,\mu_s}[-\log \hat{\mu}(y|x)]-\E_{\Dc_s,\mu_s}[-\log \mu_s(y|x)], \nonumber
    \end{align}
    where the first difference can be bounded similarly to \eqref{eq:idk4}, and the second difference is the KL divergence taken expectation on $\Dc_s$, and thus we have
    \begin{align}\label{eq:idk5}
        &\E_{\hDc,\hmu}[-\log \hat{\mu}(y|x)]-\E_{\Dc_s,\mu_s}[-\log \mu_s(y|x)] \nonumber\\
        &\leq C_p \sum_x|\hDc(x)-\Dc_s(x)|+C_p\sum_x \Dc_s(x)\sum_y|\hmu(y|x)-\mu_s(y|x)|+\E_{\Dc_s}[D_{KL}\big(\mu_s(x)|\hmu(x)\big)].
    \end{align}
    Plugging \eqref{eq:idk4}, \eqref{eq:second diff} and \eqref{eq:idk5} in \eqref{eq:idk0} gives the result.

\subsection{Proof of Theorem \ref{theorem:alignment loss constraint ft bound}}
\label{proof_theorem:alignment loss constraint ft bound}

Define $\hat{\mu}_f(\cdot|x)$ as an output distribution with $\hat{\mu}_f(y|x)=\mu_f(y|x)$ given any $x$ on the support of $\Dc_f$, and $\hat{\mu}_f(y|x)=\hat{\mu}(y|x)$ otherwise. 
    By optimality of $P_\theta$ for problem \eqref{formulation:safety loss penalty}, we first have
      \begin{align}
        \E_{\Dc_f,\mu_f}[\lpt]+\lambda \E_{\hDc,\hmu}[\lpt] 
        &\leq \E_{\Dc_f,\mu_f}[-\log \hmu_f(y|x)]+ \lambda \E_{\hDc,\hmu}[-\log \hmu_f] .
    \end{align}
   After rearranging the last inequality, we have
    \begin{align}\label{eqLidk5}
        &\E_{\Dc_f,\mu_f}[\lpt]-\E_{\Dc_f,\mu_f(y|x)}[-\log \hmu_f(y|x)] \nonumber\\
        &\leq \lambda \E_{\hDc,\hmu}[-\log \hmu_f(y|x)]-\lambda \E_{\hDc,\hmu}[\lpt]  \nonumber\\
        &= \lambda \Big(\E_{\hDc,\hmu}[-\log \hmu_f(y|x)]-\E_{\hDc,\hmu}[\lpt]+\E_{\hDc,\hmu}[-\log\hmu(y|x)]-\E_{\hDc,\hmu}[-\log\hmu(y|x)]\Big) \nonumber\\
        &\leq \lambda \Big(\E_{\hDc,\hmu}[-\log \hmu_f(y|x)]-\E_{\hDc,\hmu}[-\log\hmu(y|x)]\Big)
    \end{align}
    where the last inequality follows from $\E_{\hDc,\hmu}[-\log P_\theta(y|x)]\geq \E_{\hDc,\hmu}[-\log\hmu(y|x)]$.
    We write $\hDc\cap \Dc_f$ as shorthand notation for the intersection of the support of $\hDc$ and the the support of $\Dc_f$. Continuing from \eqref{eq:idk5}, we have
    \begin{align}
    &\E_{\Dc_f,\mu_f}[\lpt]-\E_{\Dc_f,\mu_f}[-\log \hmu_f(y|x)] \nonumber\\
        &\leq \lambda\E_{\hDc,\hmu}[-\log \hmu_f(y|x)]-\lambda\E_{\hDc,\hmu}[-\log\hmu(y|x)]\nonumber\\
        &= \lambda\sum_{x\in\hDc\cap\Dc_f}\hDc(x)\sum_y \hmu(y|x)(-\log\hmu_f(y|x))+\lambda\sum_{x\in(\hDc-(\hDc\cap\Dc_f))}\hDc(x)\sum_y \hmu(y|x)(-\log\hmu_f(y|x))\nonumber\\
        &-\lambda\sum_{x\in\hDc\cap\Dc_f}\hDc(x)\sum_y \hmu(y|x)(-\log\hmu(y|x))-\lambda\sum_{x\in\hDc\cap\Dc_f}\hDc(x)\sum_y \hmu(y|x)(-\log\hmu(y|x))\nonumber\\
        &= \lambda\sum_{x\in\hDc\cap\Dc_f}\hDc(x)\sum_y \hmu(y|x)(-\log\mu_f(y|x))+\lambda\sum_{x\in(\hDc-(\hDc\cap\Dc_f))}\hDc(x)\sum_y \hmu(y|x)(-\log\hmu(y|x))\nonumber\\
        &-\lambda\sum_{x\in\hDc\cap\Dc_f}\hDc(x)\sum_y \hmu(y|x)(-\log\hmu(y|x))-\lambda\sum_{x\in(\hDc-(\hDc\cap\Dc_f))}\hDc(x)\sum_y \hmu(y|x)(-\log\hmu(y|x))\nonumber\\
        &= \lambda \Big(\sum_{x\in\hDc\cap\Dc_f}\hDc(x)\sum_y \hmu(y|x)(-\log\mu_f(y|x))-\sum_{x\in\hDc\cap\Dc_f}\hDc(x)\sum_y \hmu(y|x)(-\log\hmu(y|x))\Big) \nonumber\\
        &=\lambda\sum_{x\in\hDc\cap \Dc_f}\hDc(x)\sum_y \hmu(y|x)\log \frac{\hmu(y|x)}{\mu_f(y|x)}\nonumber\\
        &=\lambda \sum_{x\in\hDc \cap \Dc_f}\hDc(x)D_{KL}(\hmu(x)|\mu_f(x))
    \end{align}
    where the second equality follows from the definition of $\hmu_f$.
    This completes the proof.

\subsection{Proof of Theorem \ref{theorem:alignment param constraint alignment bound}}
\label{proof_theorem:alignment param constraint alignment bound}

    The safety alignment gap can be decomposed into
    \begin{align}
        G_s(P_\theta) &= \E_{x\sim\Dc_s,y\sim\mu_s(x)}[-\log P_\theta(y|x)]-\E_{x\sim\Dc_s,y\sim\mu_s(x)}[-\log \mu_s(y|x)] \nonumber\\
        &=\E_{x\sim\Dc_s,y\sim\mu_s(x)}[-\log P_\theta(y|x)]-\E_{x\sim\Dc_s,y\sim\mu_s(x)}[-\log P_{\theta_s}(y|x)]\nonumber\\
        &+\E_{x\sim\Dc_s,y\sim\mu_s(x)}[-\log P_{\theta_s}(y|x)]-\E_{x\sim\Dc_s,y\sim\mu_s(x)}[-\log \mu_s(y|x)] \nonumber\\
        &= \E_{x\sim\Dc_s,y\sim\mu_s(x)}[-\log P_\theta(y|x)]-\E_{x\sim\Dc_s,y\sim\mu_s(x)}[-\log P_{\theta_s}(y|x)]+ G_s(P_{\theta_s}) \nonumber\\
        &\leq L_s(\theta_s,\epsilon_2)\epsilon_2+G_s(P_{\theta_s}).
    \end{align}
    where the last inequality follows from Assumption \ref{assumption:locally continuous param}.

\subsection{Proof of Theorem \ref{theorem:alignment param constraint ft bound}}
\label{proof_theorem:alignment param constraint ft bound}

Given any $\theta'$ that is feasible for \eqref{formulation:parameter constraint}, by the optimality condition of $\theta$, we have
\begin{align}\label{eq:idk6}
    &\E_{\Dc_f,\mu_f}[\lpt]-\E_{\Dc_f,\mu_f}[-\log P_{\theta_s}(y|x)] \nonumber\\
    &\leq\E_{\Dc_f,\mu_f}[\log P_{\theta'}(y|x)]-\E_{\Dc_f,\mu_f}[-\log P_{\theta_s}(y|x)]\nonumber\\
    &\leq \ip{-\E_{\Dc_f,\mu_f}[\nabla_{\theta}\log P_{\theta_s}(y|x)]}{\theta'-\theta_s}+\frac{L_f'(\theta_s,\epsilon_2)}{2}\|\theta_s-\theta'\|^2
\end{align}
where the last inequality follows from Assumption \ref{assumption:locally_smooth_param}. Let $\theta'=\theta_s+\alpha  \E_{\Dc_f,\mu_f}[\nabla_{\theta}\log P_{\theta_s}(y|x)]$ where $\alpha\leq \epsilon_2/\| \E_{\Dc_f,\mu_f}[\nabla_{\theta}\log P_{\theta_s}(y|x)]\|$ will ensure the feasibility of $\theta'$ for \eqref{formulation:parameter constraint}. Then plugging this $\theta'$ into \eqref{eq:idk6} gives
\begin{align}
     \E_{\Dc_f,\mu_f}[\lpt]-\E_{\Dc_f,\mu_f}[-\log P_{\theta_s}(y|x)] 
     \leq \big(-\alpha+\frac{L_f'(\theta_s,\epsilon_2)}{2}\alpha^2\big)\big\|\E_{\Dc_f,\mu_f}[\nabla_{\theta}\log P_{\theta_s}(y|x)]\big\|^2.\nonumber
\end{align}
Additionally choosing  $\alpha\leq 1/L_f'(\theta_s,\epsilon_2)$ gives
\begin{align}
    \E_{\Dc_f,\mu_f}[\lpt]-\E_{\Dc_f,\mu_f}[-\log P_{\theta_s}(y|x)]  \leq -\frac{\big\| \E_{\Dc_f,\mu_f}[\nabla_{\theta}\log P_{\theta_s}(y|x)]\big\|^2}{2 L_f'(\theta_s,\epsilon_2)}
\end{align}
which can be rewritten as
\begin{align}
    &\E_{\Dc_f,\mu_f}[\lpt]-\E_{\Dc_f,\mu_f}[-\log \mu_f(y|x)] \nonumber\\
    &\leq \E_{\Dc_f,\mu_f}[-\log P_{\theta_s}(y|x)]-\E_{\Dc_f,\mu_f}[-\log \mu_f(y|x)] -\frac{\big\| \E_{\Dc_f,\mu_f}[\nabla_{\theta}\log P_{\theta_s}(y|x)]\big\|^2}{2 L_f'(\theta_s,\epsilon_2)}.
\end{align}
This completes the proof.

\subsection{Additional Experimental Details}
\subsubsection{Hyper-parameters}
For the initial alignment of Llama-2-7B base model on the Orca dataset, we perform full-parameter SFT on the base model. We use Adam optimizer with an initial learning rate of $1\times10^{-5}$, a batch size of $16$, and align for 3 epochs.
Given the aligned model, we then test our capability of fine-tuning results.
We implement LoRA fine-tuning with rank $16$, $\alpha=16$ without dropout on all the query and value weight matrices in the attention layers, which results in approximately 8.4 million trainable parameters of the Llama-2-7B model. We use Adam optimizer in all experiments.
We use a batch size of $16$ and an initial learning rate of $1\times 10^{-5}$. We fine-tune the model for 3 epochs. 

\subsubsection{Loss calculation}
In all the plots, we calculate the safety alignment or capability training loss gaps following \eqref{eq:safety alignmetn gap} or \eqref{eq:capability gap}, respectively. For example, the safety alignment loss gap would be calculated as an approximation of \eqref{eq:safety alignmetn gap}:
\begin{align}
    \frac{1}{|\Dc_{orca}|}\big(\sum_{\{x,y\}\in\Dc_{orca}}[-\log P_\theta(y|x)] -\sum_{\{x,y\}\in\Dc_{orca}}[-\log P_{\theta_s}(y|x)]\big)
\end{align}
where $\Dc_{orca}$ contains $2\times 10^4$ entries of training data from the Orca dataset, and the summation can be viewed as an approximation of the expectation in \eqref{eq:safety alignmetn gap}. In addition, $ P_{\theta_s}$ is the aligned model's output distribution, which is an approximation of the Orca's output target distribution $\mu_s$. 

Similarly, for the capability loss gap, we calculate it via the following approximation of \eqref{eq:capability gap}:
\begin{align}
    \frac{1}{|\Dc_{ft}|}\big(\sum_{x,y\in\Dc_{ft}}[-\log P_\theta(y|x)] -\sum_{x,y\in\Dc_{ft}}[-\log P_{\theta_{ft}}(y|x)]\big)
\end{align}
where $\Dc_{ft}$ is the capability fine-tuning dataset specified in each experiment, where each dataset contains $2\times 10^4$ entries. And $P_{\theta_{ft}}$ is the output distribution of the model trained on the capability dataset. Thus, $P_{\theta_{ft}}$ is an approximation of $\mu_f$. %For all datasets, we used the original train and test splits.

\subsubsection{Techniques used for computational efficiency}
We used Deepspeed \cite{rasley2020deepspeed} to perform ZeRO distributed training and gradient accumulation. For ZeRO, we used ZeRO stage 3, where the optimizer states, model parameters, and training data will be split among devices if multiple GPUs are used. For gradient accumulation, we used a micro train batch size of $4$, thus with a batch size of $16$, the model update would happen every $4$ gradient accumulation step. We also used flash attention \cite{dao2023flashattention2} to speed up the attention calculations.

%\subsubsection{Code} Our code is attached for review and will be made publicly available.

\backmatter

\bmhead{Acknowledgements}

This work was supported by IBM through the IBM-Rensselaer Future of Computing Research Collaboration, and the National
Science Foundation Project 2401297, and 2412486.

\bibliography{bob}% common bib file

%% BioMed_Central_Bib_Style_v1.01

\begin{thebibliography}{19}
% BibTex style file: bmc-mathphys.bst (version 2.1), 2014-07-24
\ifx \bisbn   \undefined \def \bisbn  #1{ISBN #1}\fi
\ifx \binits  \undefined \def \binits#1{#1}\fi
\ifx \bauthor  \undefined \def \bauthor#1{#1}\fi
\ifx \batitle  \undefined \def \batitle#1{#1}\fi
\ifx \bjtitle  \undefined \def \bjtitle#1{#1}\fi
\ifx \bvolume  \undefined \def \bvolume#1{\textbf{#1}}\fi
\ifx \byear  \undefined \def \byear#1{#1}\fi
\ifx \bissue  \undefined \def \bissue#1{#1}\fi
\ifx \bfpage  \undefined \def \bfpage#1{#1}\fi
\ifx \blpage  \undefined \def \blpage #1{#1}\fi
\ifx \burl  \undefined \def \burl#1{\textsf{#1}}\fi
\ifx \doiurl  \undefined \def \doiurl#1{\url{https://doi.org/#1}}\fi
\ifx \betal  \undefined \def \betal{\textit{et al.}}\fi
\ifx \binstitute  \undefined \def \binstitute#1{#1}\fi
\ifx \binstitutionaled  \undefined \def \binstitutionaled#1{#1}\fi
\ifx \bctitle  \undefined \def \bctitle#1{#1}\fi
\ifx \beditor  \undefined \def \beditor#1{#1}\fi
\ifx \bpublisher  \undefined \def \bpublisher#1{#1}\fi
\ifx \bbtitle  \undefined \def \bbtitle#1{#1}\fi
\ifx \bedition  \undefined \def \bedition#1{#1}\fi
\ifx \bseriesno  \undefined \def \bseriesno#1{#1}\fi
\ifx \blocation  \undefined \def \blocation#1{#1}\fi
\ifx \bsertitle  \undefined \def \bsertitle#1{#1}\fi
\ifx \bsnm \undefined \def \bsnm#1{#1}\fi
\ifx \bsuffix \undefined \def \bsuffix#1{#1}\fi
\ifx \bparticle \undefined \def \bparticle#1{#1}\fi
\ifx \barticle \undefined \def \barticle#1{#1}\fi
\bibcommenthead
\ifx \bconfdate \undefined \def \bconfdate #1{#1}\fi
\ifx \botherref \undefined \def \botherref #1{#1}\fi
\ifx \url \undefined \def \url#1{\textsf{#1}}\fi
\ifx \bchapter \undefined \def \bchapter#1{#1}\fi
\ifx \bbook \undefined \def \bbook#1{#1}\fi
\ifx \bcomment \undefined \def \bcomment#1{#1}\fi
\ifx \oauthor \undefined \def \oauthor#1{#1}\fi
\ifx \citeauthoryear \undefined \def \citeauthoryear#1{#1}\fi
\ifx \endbibitem  \undefined \def \endbibitem {}\fi
\ifx \bconflocation  \undefined \def \bconflocation#1{#1}\fi
\ifx \arxivurl  \undefined \def \arxivurl#1{\textsf{#1}}\fi
\csname PreBibitemsHook\endcsname

%%% 1
\bibitem[\protect\citeauthoryear{Ouyang et~al.}{2022}]{ouyang2022training}
\begin{barticle}
\bauthor{\bsnm{Ouyang}, \binits{L.}},
\bauthor{\bsnm{Wu}, \binits{J.}},
\bauthor{\bsnm{Jiang}, \binits{X.}},
\bauthor{\bsnm{Almeida}, \binits{D.}},
\bauthor{\bsnm{Wainwright}, \binits{C.}},
\bauthor{\bsnm{Mishkin}, \binits{P.}},
\bauthor{\bsnm{Zhang}, \binits{C.}},
\bauthor{\bsnm{Agarwal}, \binits{S.}},
\bauthor{\bsnm{Slama}, \binits{K.}},
\bauthor{\bsnm{Ray}, \binits{A.}}, \betal:
\batitle{Training language models to follow instructions with human feedback}.
\bjtitle{Advances in neural information processing systems}
\bvolume{35},
\bfpage{27730}--\blpage{27744}
(\byear{2022})
\end{barticle}
\endbibitem

%%% 2
\bibitem[\protect\citeauthoryear{Ding et~al.}{2023}]{ding2023parameter}
\begin{barticle}
\bauthor{\bsnm{Ding}, \binits{N.}},
\bauthor{\bsnm{Qin}, \binits{Y.}},
\bauthor{\bsnm{Yang}, \binits{G.}},
\bauthor{\bsnm{Wei}, \binits{F.}},
\bauthor{\bsnm{Yang}, \binits{Z.}},
\bauthor{\bsnm{Su}, \binits{Y.}},
\bauthor{\bsnm{Hu}, \binits{S.}},
\bauthor{\bsnm{Chen}, \binits{Y.}},
\bauthor{\bsnm{Chan}, \binits{C.-M.}},
\bauthor{\bsnm{Chen}, \binits{W.}}, \betal:
\batitle{Parameter-efficient fine-tuning of large-scale pre-trained language models}.
\bjtitle{Nature Machine Intelligence}
\bvolume{5}(\bissue{3}),
\bfpage{220}--\blpage{235}
(\byear{2023})
\end{barticle}
\endbibitem

%%% 3
\bibitem[\protect\citeauthoryear{Qi et~al.}{2024}]{qi2024finetuning}
\begin{bchapter}
\bauthor{\bsnm{Qi}, \binits{X.}},
\bauthor{\bsnm{Zeng}, \binits{Y.}},
\bauthor{\bsnm{Xie}, \binits{T.}},
\bauthor{\bsnm{Chen}, \binits{P.-Y.}},
\bauthor{\bsnm{Jia}, \binits{R.}},
\bauthor{\bsnm{Mittal}, \binits{P.}},
\bauthor{\bsnm{Henderson}, \binits{P.}}:
\bctitle{Fine-tuning aligned language models compromises safety, even when users do not intend to!}
In: \bbtitle{International Conference on Learning Representations}
(\byear{2024})
\end{bchapter}
\endbibitem

%%% 4
\bibitem[\protect\citeauthoryear{Chen}{2025}]{chen2025computational}
\begin{botherref}
\oauthor{\bsnm{Chen}, \binits{P.-Y.}}:
Computational safety for generative {AI}: A signal processing perspective.
arXiv preprint arXiv:2502.12445
(2025)
\end{botherref}
\endbibitem

%%% 5
\bibitem[\protect\citeauthoryear{Choudhury et~al.}{2025}]{choudhury2025promise}
\begin{botherref}
\oauthor{\bsnm{Choudhury}, \binits{M.}},
\oauthor{\bsnm{Elyoseph}, \binits{Z.}},
\oauthor{\bsnm{Fast}, \binits{N.J.}},
\oauthor{\bsnm{Ong}, \binits{D.C.}},
\oauthor{\bsnm{Nsoesie}, \binits{E.O.}},
\oauthor{\bsnm{Pavlick}, \binits{E.}}:
The promise and pitfalls of generative ai.
Nature Reviews Psychology,
1--6
(2025)
\end{botherref}
\endbibitem

%%% 6
\bibitem[\protect\citeauthoryear{Chang et~al.}{2025}]{chang2025red}
\begin{barticle}
\bauthor{\bsnm{Chang}, \binits{C.T.}},
\bauthor{\bsnm{Farah}, \binits{H.}},
\bauthor{\bsnm{Gui}, \binits{H.}},
\bauthor{\bsnm{Rezaei}, \binits{S.J.}},
\bauthor{\bsnm{Bou-Khalil}, \binits{C.}},
\bauthor{\bsnm{Park}, \binits{Y.-J.}},
\bauthor{\bsnm{Swaminathan}, \binits{A.}},
\bauthor{\bsnm{Omiye}, \binits{J.A.}},
\bauthor{\bsnm{Kolluri}, \binits{A.}},
\bauthor{\bsnm{Chaurasia}, \binits{A.}}, \betal:
\batitle{Red teaming chatgpt in medicine to yield real-world insights on model behavior}.
\bjtitle{npj Digital Medicine}
\bvolume{8}(\bissue{1}),
\bfpage{149}
(\byear{2025})
\end{barticle}
\endbibitem

%%% 7
\bibitem[\protect\citeauthoryear{Huang et~al.}{2024}]{huang2024harmful}
\begin{botherref}
\oauthor{\bsnm{Huang}, \binits{T.}},
\oauthor{\bsnm{Hu}, \binits{S.}},
\oauthor{\bsnm{Ilhan}, \binits{F.}},
\oauthor{\bsnm{Tekin}, \binits{S.F.}},
\oauthor{\bsnm{Liu}, \binits{L.}}:
Harmful fine-tuning attacks and defenses for large language models: A survey.
arXiv preprint arXiv:2409.18169
(2024)
\end{botherref}
\endbibitem

%%% 8
\bibitem[\protect\citeauthoryear{Bianchi et~al.}{2024}]{bianchi2024safetytuned}
\begin{bchapter}
\bauthor{\bsnm{Bianchi}, \binits{F.}},
\bauthor{\bsnm{Suzgun}, \binits{M.}},
\bauthor{\bsnm{Attanasio}, \binits{G.}},
\bauthor{\bsnm{Rottger}, \binits{P.}},
\bauthor{\bsnm{Jurafsky}, \binits{D.}},
\bauthor{\bsnm{Hashimoto}, \binits{T.}},
\bauthor{\bsnm{Zou}, \binits{J.}}:
\bctitle{Safety-tuned {LL}a{MA}s: Lessons from improving the safety of large language models that follow instructions}.
In: \bbtitle{International Conference on Learning Representations}
(\byear{2024})
\end{bchapter}
\endbibitem

%%% 9
\bibitem[\protect\citeauthoryear{Shen et~al.}{2025}]{shen2025seal}
\begin{bchapter}
\bauthor{\bsnm{Shen}, \binits{H.}},
\bauthor{\bsnm{Chen}, \binits{P.-Y.}},
\bauthor{\bsnm{Das}, \binits{P.}},
\bauthor{\bsnm{Chen}, \binits{T.}}:
\bctitle{{SEAL}: Safety-enhanced aligned {LLM} fine-tuning via bilevel data selection}.
In: \bbtitle{International Conference on Learning Representations}
(\byear{2025})
\end{bchapter}
\endbibitem

%%% 10
\bibitem[\protect\citeauthoryear{Peng et~al.}{2024}]{peng2024navigating}
\begin{bchapter}
\bauthor{\bsnm{Peng}, \binits{S.}},
\bauthor{\bsnm{Chen}, \binits{P.-Y.}},
\bauthor{\bsnm{Hull}, \binits{M.D.}},
\bauthor{\bsnm{Chau}, \binits{D.H.}}:
\bctitle{Navigating the safety landscape: Measuring risks in finetuning large language models}.
In: \bbtitle{Neural Information Processing Systems}
(\byear{2024})
\end{bchapter}
\endbibitem

%%% 11
\bibitem[\protect\citeauthoryear{Hsu et~al.}{2024}]{hsu2024safe}
\begin{bchapter}
\bauthor{\bsnm{Hsu}, \binits{C.-Y.}},
\bauthor{\bsnm{Tsai}, \binits{Y.-L.}},
\bauthor{\bsnm{Lin}, \binits{C.-H.}},
\bauthor{\bsnm{Chen}, \binits{P.-Y.}},
\bauthor{\bsnm{Yu}, \binits{C.-M.}},
\bauthor{\bsnm{Huang}, \binits{C.-Y.}}:
\bctitle{Safe {LoRA}: The silver lining of reducing safety risks when finetuning large language models}.
In: \bbtitle{Neural Information Processing Systems}
(\byear{2024})
\end{bchapter}
\endbibitem

%%% 12
\bibitem[\protect\citeauthoryear{Touvron et~al.}{2023}]{touvron2023llama}
\begin{botherref}
\oauthor{\bsnm{Touvron}, \binits{H.}},
\oauthor{\bsnm{Martin}, \binits{L.}},
\oauthor{\bsnm{Stone}, \binits{K.}},
\oauthor{\bsnm{Albert}, \binits{P.}},
\oauthor{\bsnm{Almahairi}, \binits{A.}},
\oauthor{\bsnm{Babaei}, \binits{Y.}},
\oauthor{\bsnm{Bashlykov}, \binits{N.}},
\oauthor{\bsnm{Batra}, \binits{S.}},
\oauthor{\bsnm{Bhargava}, \binits{P.}},
\oauthor{\bsnm{Bhosale}, \binits{S.}}, et al.:
Llama 2: Open foundation and fine-tuned chat models.
arXiv preprint arXiv:2307.09288
(2023)
\end{botherref}
\endbibitem

%%% 13
\bibitem[\protect\citeauthoryear{Lian et~al.}{2023}]{OpenOrca}
\begin{botherref}
\oauthor{\bsnm{Lian}, \binits{W.}},
\oauthor{\bsnm{Goodson}, \binits{B.}},
\oauthor{\bsnm{Pentland}, \binits{E.}},
\oauthor{\bsnm{Cook}, \binits{A.}},
\oauthor{\bsnm{Vong}, \binits{C.}},
\oauthor{\bsnm{"Teknium"}}:
OpenOrca: An Open Dataset of GPT Augmented FLAN Reasoning Traces.
HuggingFace
(2023)
\end{botherref}
\endbibitem

%%% 14
\bibitem[\protect\citeauthoryear{Achiam et~al.}{2023}]{achiam2023gpt}
\begin{botherref}
\oauthor{\bsnm{Achiam}, \binits{J.}},
\oauthor{\bsnm{Adler}, \binits{S.}},
\oauthor{\bsnm{Agarwal}, \binits{S.}},
\oauthor{\bsnm{Ahmad}, \binits{L.}},
\oauthor{\bsnm{Akkaya}, \binits{I.}},
\oauthor{\bsnm{Aleman}, \binits{F.L.}},
\oauthor{\bsnm{Almeida}, \binits{D.}},
\oauthor{\bsnm{Altenschmidt}, \binits{J.}},
\oauthor{\bsnm{Altman}, \binits{S.}},
\oauthor{\bsnm{Anadkat}, \binits{S.}}, et al.:
{GPT}-4 technical report.
arXiv preprint arXiv:2303.08774
(2023)
\end{botherref}
\endbibitem

%%% 15
\bibitem[\protect\citeauthoryear{Taori et~al.}{2023}]{alpaca}
\begin{botherref}
\oauthor{\bsnm{Taori}, \binits{R.}},
\oauthor{\bsnm{Gulrajani}, \binits{I.}},
\oauthor{\bsnm{Zhang}, \binits{T.}},
\oauthor{\bsnm{Dubois}, \binits{Y.}},
\oauthor{\bsnm{Li}, \binits{X.}},
\oauthor{\bsnm{Guestrin}, \binits{C.}},
\oauthor{\bsnm{Liang}, \binits{P.}},
\oauthor{\bsnm{Hashimoto}, \binits{T.B.}}:
Stanford {Alpaca}: An Instruction-following {LLaMA} model.
GitHub
(2023)
\end{botherref}
\endbibitem

%%% 16
\bibitem[\protect\citeauthoryear{Muennighoff et~al.}{2023}]{muennighoff2023octopack}
\begin{bchapter}
\bauthor{\bsnm{Muennighoff}, \binits{N.}},
\bauthor{\bsnm{Liu}, \binits{Q.}},
\bauthor{\bsnm{Zebaze}, \binits{A.}},
\bauthor{\bsnm{Zheng}, \binits{Q.}},
\bauthor{\bsnm{Hui}, \binits{B.}},
\bauthor{\bsnm{Zhuo}, \binits{T.Y.}},
\bauthor{\bsnm{Singh}, \binits{S.}},
\bauthor{\bsnm{Tang}, \binits{X.}},
\bauthor{\bsnm{Werra}, \binits{L.V.}},
\bauthor{\bsnm{Longpre}, \binits{S.}}:
\bctitle{{OctoPack}: Instruction tuning code large language models}.
In: \bbtitle{NeurIPS 2023 Workshop on Instruction Tuning and Instruction Following}
(\byear{2023})
\end{bchapter}
\endbibitem

%%% 17
\bibitem[\protect\citeauthoryear{Lee et~al.}{2023}]{platypus2023}
\begin{bchapter}
\bauthor{\bsnm{Lee}, \binits{A.N.}},
\bauthor{\bsnm{Hunter}, \binits{C.J.}},
\bauthor{\bsnm{Ruiz}, \binits{N.}}:
\bctitle{Platypus: Quick, cheap, and powerful refinement of {LLMs}}.
In: \bbtitle{NeurIPS 2023 Workshop on Instruction Tuning and Instruction Following}
(\byear{2023})
\end{bchapter}
\endbibitem

%%% 18
\bibitem[\protect\citeauthoryear{Rasley et~al.}{2020}]{rasley2020deepspeed}
\begin{bchapter}
\bauthor{\bsnm{Rasley}, \binits{J.}},
\bauthor{\bsnm{Rajbhandari}, \binits{S.}},
\bauthor{\bsnm{Ruwase}, \binits{O.}},
\bauthor{\bsnm{He}, \binits{Y.}}:
\bctitle{Deepspeed: System optimizations enable training deep learning models with over 100 billion parameters}.
In: \bbtitle{Proceedings of the 26th ACM SIGKDD International Conference on Knowledge Discovery \& Data Mining},
pp. \bfpage{3505}--\blpage{3506}
(\byear{2020})
\end{bchapter}
\endbibitem

%%% 19
\bibitem[\protect\citeauthoryear{Dao}{2024}]{dao2023flashattention2}
\begin{bchapter}
\bauthor{\bsnm{Dao}, \binits{T.}}:
\bctitle{Flash{A}ttention-2: Faster attention with better parallelism and work partitioning}.
In: \bbtitle{International Conference on Learning Representations}
(\byear{2024})
\end{bchapter}
\endbibitem

\end{thebibliography}
%% if required, the content of .bbl file can be included here once bbl is generated
%%\input sn-article.bbl

\end{document}